\documentclass[conference]{IEEEtran}
\IEEEoverridecommandlockouts
\usepackage{cite}
\usepackage{amsmath,amssymb,amsfonts}
\usepackage{graphicx}
\usepackage{textcomp}
\usepackage{xcolor}
\def\BibTeX{{\rm B\kern-.05em{\sc i\kern-.025em b}\kern-.08em
    T\kern-.1667em\lower.7ex\hbox{E}\kern-.125emX}}

\usepackage{multirow}
\usepackage{amsmath}
\usepackage{amsthm}
\usepackage{graphicx}
\usepackage{subcaption}
\usepackage{amsfonts}

\usepackage{mathtools} 
\usepackage{alltt, xspace, times, epsfig, url} 
\usepackage{gensymb,xfrac,array,booktabs,bm,xcolor} 
\usepackage[linesnumbered,lined,ruled]{algorithm2e}

\newtheorem{lemma}{Lemma}
\newtheorem{definition}{Definition}
\newtheorem{prop}{Proposition}

\makeatletter
\newcommand{\nosemic}{\renewcommand{\@endalgocfline}{\relax}}
\newcommand{\dosemic}{\renewcommand{\@endalgocfline}{\algocf@endline}}
\let\oldnl\nl
\newcommand{\nonl}{\renewcommand{\nl}{\let\nl\oldnl}}

\begin{document}


\title{Gradient-Leakage Resilient Federated Learning}

 \author{\IEEEauthorblockN{\mbox{Wenqi Wei\IEEEauthorrefmark{2}, Ling Liu\IEEEauthorrefmark{2}, Yanzhao Wu\IEEEauthorrefmark{2}, Gong Su\IEEEauthorrefmark{1}, Arun Iyengar\IEEEauthorrefmark{1}}} 
\IEEEauthorblockA{\IEEEauthorrefmark{2}
Georgia Institute of Technology, School of CS,
Atlanta, GA 30332, USA \\ } 
 \IEEEauthorblockA{\IEEEauthorrefmark{1} IBM T. J. Watson Research Center, Yorktown Heights, NY 10598, USA}
}

\maketitle

\begin{abstract}
Federated learning(FL) is an emerging distributed learning paradigm with default client privacy because clients can keep sensitive data on their devices and only share local training parameter updates with the federated server. However, recent studies reveal that gradient leakages in FL may compromise the privacy of client training data. This paper presents a gradient leakage resilient approach to privacy-preserving federated learning with per training example-based client differential privacy, coined as Fed-CDP. It makes three original contributions. First, we identify three types of client gradient leakage threats in federated learning even with encrypted client-server communications. We articulate when and why the conventional server coordinated differential privacy approach, coined as Fed-SDP, is insufficient to protect the privacy of the training data. Second, we introduce Fed-CDP, the per example-based client differential privacy algorithm, and provide a formal analysis of Fed-CDP with the $(\epsilon, \delta)$ differential privacy guarantee, and a formal comparison between Fed-CDP and Fed-SDP in terms of privacy accounting. Third, we formally analyze the privacy-utility trade-off for providing differential privacy guarantee by Fed-CDP and present a dynamic decay noise-injection policy to further improve the accuracy and resiliency of Fed-CDP. We evaluate and compare Fed-CDP and Fed-CDP(decay) with Fed-SDP in terms of differential privacy guarantee and gradient leakage resilience over five benchmark datasets. The results show that the Fed-CDP approach outperforms conventional Fed-SDP in terms of resilience to client gradient leakages while offering competitive accuracy performance in federated learning.
\end{abstract}

\begin{IEEEkeywords}
Distributed systems, federated learning, differential privacy, gradient leakage, attack resilience.
\end{IEEEkeywords}

\section{Introduction}
Federated learning (FL) decouples the learning task from the centralized server to a distributed collection of client nodes~\cite{vanhaesebrouck2016decentralized,mcmahan2017communication,konevcny2016federated,bonawitz2019towards,zhao2018federated}. It collaboratively learns a shared global model over the distributed collection of training data residing on end-user devices of the participating clients. Each client maintains its private data locally by downloading the global model update from the server at each round of the federated learning, performing local training, and sharing its local model update via encrypted communication to the server. 
 

However, the distributed nature and the default privacy in federated learning are insufficient for protecting client training data from gradient leakage attacks. Recent studies~\cite{zhu2019deep,geiping2020inverting,wei2020framework} reveal that if an adversary intercepts the local gradient update of a client before the server performs the federated aggregation to generate the global parameter update for the next round of federated learning, the adversary can steal the sensitive local training data of this client using the leaked gradients by simply performing reconstruction attack~\cite{wei2020framework}, or membership and attribute inference attacks \cite{nasr2018comprehensive,truex2018towards,shokri2017membership}.


To circumvent such gradient leakage threats, we present Fed-CDP, the per example-based client differential privacy approach with three original contributions.
First, we categorize client gradient leakage threats into three types based on whether the gradient information was leaked at the server, the client, or the training example level. We analyze when and why the conventional server coordinated differential privacy approach~\cite{mcmahan2017learning,geyer2017differentially}, coined as Fed-SDP in this paper, is insufficient to protect the privacy of client training data in the presence of instance-level gradient leakage attacks. 
Second, we introduce Fed-CDP to provide the formal instance-level and client-level differential privacy guarantee. In contrast to Fed-SDP, which adds noise to the shared gradient update by a client at each round of the FL joint training, Fed-CDP adds noise to per example at each iteration per client.
The differential privacy spending is computed by the privacy accounting using the moments accountant~\cite{abadi2016deep} for both Fed-SDP and Fed-CDP. 
Finally, we provide a formal analysis of the privacy-utility trade-off for providing differential privacy guarantee by Fed-CDP, with the maximal distortion bound without deteriorating the performance of federated learning. Guided by this formal trade-off analysis, we optimize Fed-CDP by enabling a dynamic decay noise-injection policy, which boosts both the accuracy performance and the resilience against gradient leakage attacks. Evaluated on five benchmark datasets, we show that per-example based Fed-CDP and Fed-CDP (decay) outperforms per client based Fed-SDP in terms of accuracy performance, differential privacy guarantee, and gradient leakage resilience.

\begin{figure*}[ht]
\begin{minipage}[t]{0.40\linewidth}
 \centerline{\includegraphics[scale=.51]{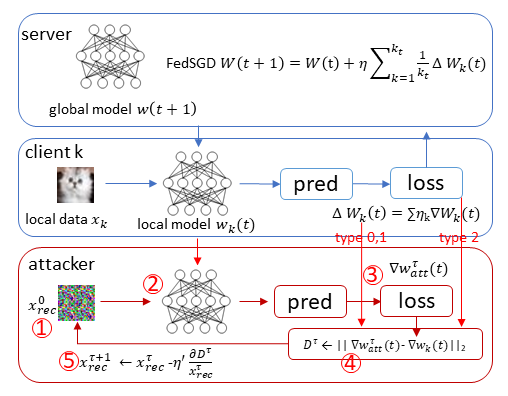}}
 \subcaption{\small Attack schema}
 \label{fig:attack_schema}
\end{minipage}
\begin{minipage}[t]{0.59\linewidth}
 \centerline{\includegraphics[scale=.48]{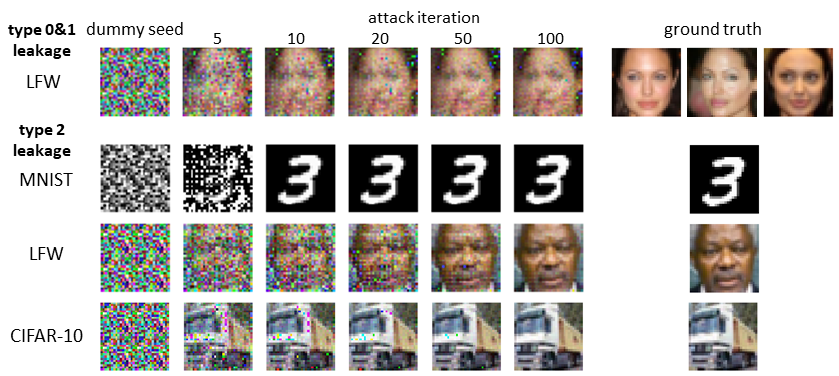}}
\subcaption{\small  Attack illustration.}
 \label{fig:attack_example_type2}
\end{minipage}
\caption{\small Gradient leakage attacks in Federated Learning. Type-0\&1 leakage attack by batched update and type-2 attacks by example}
 \label{fig:attack}
 \vspace{-0.4cm}
\end{figure*}

\section{Related Work}

Gradient leakage threats refer to unauthorized disclosure of private client training data based on the leaked gradient information and are one of the dominating privacy leakage threats in both centralized and distributed deep learning systems. 
In centralized deep learning, membership inference~\cite{shokri2017membership,melis2019exploiting,truex2018towards,nasr2018comprehensive} is known to expose the private training data by exploiting the statistical differences between the model prediction on the data possibly seen during training and the prediction on unseen data. 
\cite{song2017machine}
performed minor modifications to training algorithms to have them memorize a significant amount
of information about the training data. 
\cite{aono2017privacy} inferred private data based on the proportionality between the training data and the gradient updates but fails to be generalized to convolutional neural networks due to the relative size of convolution weights and data features. 
\cite{fredrikson2015model} proposed the model inversion attack to expose privately trained model parameters by exploiting confidence values revealed from legitimate access to the prediction API. 
GAN-based reconstruction attack~\cite{hitaj2017deep,wang2019beyond} utilizes generative adversarial networks to learn private training data based on the gradient information and the confidence of prediction results. 
In distributed federated learning,  \cite{geiping2020inverting,zhu2019deep,wei2020framework} showed the vulnerabilities of client gradient leakage attacks.
\cite{wei2020framework} showed that such gradient-based client privacy leakage (CPL) attacks display high reconstruction success rate with high attack effect and low attack cost. 

Multiple attempts have been made to offer privacy-preserving federated learning. Cryptographic approaches~\cite{shi2011privacy} and secure multi-party computation (SMP) \cite{bonawitz2017practical,chase2017private} ensure secure communication of shared parameter updates between a client and its federated server. However, it does not secure the client data prior to  encryption for transport or after decryption for the server aggregation.
\cite{papernot2018scalable} proposed PATE, a teacher-student ensemble, which relies on a trusted server for private aggregation, but the involved third party can bring new risk to the FL system. 
\cite{shokri2015privacy} proposed a distributed learning with selective and random sharing of model parameters, offering the shared model parameter level privacy. In recent proposals for differentially private federated learning~\cite{mcmahan2017learning,geyer2017differentially},   the noise is added to the per-client shared gradient updates at each round of FL, the effect and cost of such differential privacy guarantee are the same regardless of whether the clipping and the noise injection are done at the server or at each client. In this paper, we coin this conventional approach as Fed-SDP. We will show that both the selective parameter sharing approach and Fed-SDP approach fail to safeguard the client’s private training data in the presence of instance-level gradient leakage attacks.  

\section{Gradient Leakage Threats}
\label{sec3}

\textbf{Threat Model.} Our threat model adopts the two common assumptions: (i) the message communicated between a client and its FL server is encrypted; and (ii) the clients faithfully follow the rules of joint training when subscribed to a federated learning task, although the server may conduct unauthorized inference on per-client local model updates either curiously or maliciously and a client may read the local training result during or upon completion of the local training at each round of federated learning. We categorize the gradient leakage threats into three types based on the location and whether the gradient leakage is performed on per-example gradient update during local training or per client gradient update after completing the local training at each round of FL. 
The \textbf{type-0} refers to the gradient leakage attacks performed at the server by intercepting per-client shared gradient updates. The \textbf{type-1} refers to the gradient leakage attacks performed at a client on its gradient updates resulting from the completion of the local training. 
The third type of gradient leakage attacks are performed on each training example during the local training at a client and coined as the \textbf{type-2} leakage. We consider the gradient leakage attack that uses gradient-based reconstruction attack methods~\cite{zhu2019deep,geiping2020inverting,wei2020framework} to derive and steal the private client training data. Although all three types of gradient leakages are inferred by adversarial reconstruction, the first two types of leakages are launched by an unauthorized read of per-client gradient update resulting from the completion of the local training over multiple iterations at a client. The \textbf{type-2} leakages are launched by an unauthorized read of the per-example gradient updates during the local training on a client. 

Figure~\ref{fig:attack_schema} gives a sketch of the gradient leakage attack algorithm, which configures and executes the reconstruction attack in five steps: 
(1) It configures the initialization seed ($x^{0}_{rec}(t)$), a dummy data of the same resolution (or attribute structure for text) as the training data. \cite{wei2020framework} showed some significant impact of 
different initialization seeds on the attack success rate and attack cost ($\#$ attack iterations to succeed). (2) The dummy attack seed is fed into the client local model. (3) The gradient of the dummy attack seed is obtained by backpropagation. (4) The gradient loss is computed using a vector distance loss function, e.g., $L_2$, between the gradient of the attack seed and the actual gradient from the client local training. The choice of this reconstruction loss function is another tunable attack parameter. (5) The dummy attack seed is modified by the attack reconstruction learning algorithm. It aims to minimize the vector distance loss by a loss optimizer such that the gradients of the reconstructed seed $x^{i}_{rec}(t)$ at round $i$ will be closer to the actual gradient updates stolen from the client upon the completion (type 0\&1) or during the local training (type 2). This attack reconstruction learning iterates until it reaches the attack termination condition ($\mathbb{T}$), typically defined by the $\#$rounds, e.g., 300 (also a configurable attack parameter). If the reconstruction loss is smaller than the specified distance threshold then the reconstruction attack is successful. 
Figure~\ref{fig:attack_example_type2} provides a visualization by examples from three datasets. The type-0\&1 gradient leakage attack is performed on the batched gradients with the batch size of 3 from LFW dataset~\cite{huang2008labeled}. The type-2 leakage is performed on per example gradient during local train with three examples from MNIST~\cite{lecun1998mnist}, LFW and CIFAR10~\cite{krizhevsky2009learning} respectively. 
Although all three types of gradient leakage attacks succeed at the 50$^{th}$ attack iteration for these examples with $\mathbb{T}$ set to 300, type 2 leakage attacks can succeed with smaller $\#$iterations and higher attack quality for the same $\#$iterations (e.g., 100).
All the experiments on gradient leakages in this paper use the patterned random seed initialization\footnote{\url{https://github.com/git-disl/ESORICS20-CPL}}, with a $L_2$ based loss function and L-BFGS optimizer, for high attack success rate (ASR) and fast attack convergence. 





\section{Solution Approach}

\label{sec4}
We first define the core differential privacy concepts following the state-of-the-art literature. Then we introduce the reference model for federated learning (both non-private and with different privacy) and present the problem statement, followed by our proposed Fed-CDP, the per-example based client differential privacy approach to privacy-preserving federated learning, including algorithm, privacy parameters, and configuration trade-offs.

\subsection{Preliminary Concepts and Reference Model}

\begin{definition}
\textbf{Differential privacy~\cite{dwork2014algorithmic}}: a randomized mechanism $\mathcal{M}$: $\mathcal{D}\rightarrow \mathcal{R}$,  with a domain $\mathcal{D}$ of possible training data and range $\mathcal{R}$ of all possible trained models, satisfies ($\epsilon,\delta$)-differential privacy if for any two input sets $A\in \mathcal{D}$ and $A'\in \mathcal{D}$, differing with only a single example input and for any subset of outputs $B\in \mathcal{R}$, Equation~\ref{equa:dp} holds.
\begin{equation}
\Pr(\mathcal{M}(A) \in \mathcal{R}) \le e^{\epsilon}
\Pr(\mathcal{M}(A') \in \mathcal{R}) + \delta.
\label{equa:dp}
\end{equation}
\end{definition}

\begin{definition}
\textbf{Gaussian mechanism~\cite{dwork2014algorithmic}}: when $0<\epsilon<1$, applying Gaussian noise $\mathcal{N}(0, \sigma^2 S^2)$ calibrated to a real valued function: $f: \mathcal{D}\rightarrow \mathcal{R}$ such that $\mathcal{M}(A) = f(A) + \mathcal{N}(0, \sigma'^2)$, $\mathcal{M}(A)$ is $(\epsilon,\delta)$-differentially private if $\sigma {'^2} > \frac{{2\log (1.25/\delta ) \cdot {S^2}}}{{{\varepsilon ^2}}}$, where for any two adjacent inputs $A,A' \in D$, $||A-A'||_1=1$, and $S = \max \nolimits_{A,A' \in D} ||f(A)-f(A')||_2$, i.e., the maximum $l_2$ distance between $f(A)$ and $f(A')$.
\end{definition}

With Definition~2, it is straightforward to get Lemma~\ref{lemma:gaussian_mechanism}.

\begin{lemma}
Let $\sigma'^2$ in Gaussian mechanism be $\sigma^2S^2$. We have the noise scale $\sigma$ satisfying $\sigma^2>\frac{{2\log (1.25/\delta )}}{{{\varepsilon ^2}}}.$
\label{lemma:gaussian_mechanism}
\end{lemma}

\begin{definition}
\textbf{Privacy amplification~\cite{wang2019subsampled}}: Given dataset $\mathcal{D}$ with $N$ data points, define subsampling as random sampling with replacement and sampling rate as $q=n/N$ where $n$ is the sample size. If $\mathcal{M}$ is  $(\epsilon,\delta)$-differentially private, then the subsampled mechanism with sampling rate $q$ is $(\log(1+q(\exp(\epsilon)-1)),q\delta)$-differentially private.
\end{definition}

\begin{definition}
\textbf{Composition theorem~\cite{dwork2014algorithmic}}: Given a fixed $\sigma$ and $\epsilon$, then a set of consecutive inquiries of ($\epsilon,\delta$) differentially private algorithms is ($\sum \epsilon, \sum \delta$) differentially private.
\end{definition}


\begin{definition}
\textbf{Moments accountant~\cite{abadi2016deep}}: For a mechanism $\mathcal{M}$, given dataset $\mathcal{D}$ with a total of $N$ data points, given the random sampling with replacement and the sampling rate $q = n/N$ where $n$ is the sample size, and the number of iterations $T$, if $q<\frac{1}{16\sigma}$, there exist constants $c1$ and $c2$ so that for any $\epsilon  < c1q^2T$, $\mathcal{M}$ is $(\epsilon,\delta)$-differentially private for $\delta>0$ and 
\begin{equation}
    \epsilon \geq c2\frac{q\sqrt{T\log(1/\delta)}}{\sigma}
    \label{equa:momentsaccountant}
\end{equation}.
    \vspace{-0.5cm}
\end{definition}
\noindent Moments accountant is implementated\footnote{\url{https://github.com/tensorflow/privacy/blob/master/tensorflow_privacy/privacy/analysis/compute_dp_sgd_privacy.py}} in a way that privacy spending $\epsilon$ is computed when $T$, $\sigma$, $\delta$, and $q$ are given.


{\bf Reference model for federated learning.\/} Federated learning can be viewed as a publish-subscribe model, in which the server publishes its federated learning task, described by the data structure and data modality and the neural network model (algorithm) for joint training, including the important hyperparameters: total \#rounds $T$ of joint training, local batch size $B$, learning rate $\eta$, local \# iterations $L$, format for messaging between client and server, and the public key of the server for encryption. By subscribing to a published FL task, the clients $i$ informs the server its local training dataset size $N_i$ and commits to download the joint training model with the default hyperparameters, then faithfully follow the FL protocol: at each round $t \leq T$, a client downloads the current global model parameters $W(t)$ from the server, performs local model training on the private client training data over $L$ local iterations, i.e., $W_i(t)_{l+1}=W_i(t)_{l}-\eta \nabla {W_i(t)_l}, 1<l<L-1$ where $\nabla {W}$ is the gradient of the trainable network parameters and $1\leq L \leq \lceil N_i/B \rceil$.
Upon completion of the local model training at round $t$, this client $i$ sends the local training parameter updates to the server. 

Upon receiving local training parameter updates from $K_t$ out of a total of $K$ clients ($K_t < K$), i.e., $\Delta W_i(t)= W_i(t)_L- W_i(t)$ ($i=1\dots K_t$), the server aggregates the local updates from all $K_t$ clients to produce the global model parameter updates and start the next round $t+1 \leq T$ of joint training until $T$ rounds are reached. As stated in ~\cite{mcmahan2017communication}, to deal with the unstable client availability, only a small subset of $K_t$ clients are selected ($K_t < K$) to participate in each round $t$ of federated learning. Two aggregation methods are commonly used:  \textbf{FedSGD}: $W(t + 1) = W(t) + \frac{1}{K_t}\sum\nolimits_{k = 1}^{{K_t}} \Delta W_{k}(t)$ if the clients share the local model parameter updates~\cite{lin2018deep,liu2020accelerating,yao2019federated}; and  \textbf{FedAveraging}: $W(t + 1) = \sum\nolimits_{k = 1}^{{K_t}} \frac{1}{K_t}  {W_k}(t)_L$ if the clients send the locally updated model to the server~\cite{mcmahan2017communication,kamp2018efficient,ma2015adding}. With a large client population ($K$) and a small subset $K_t << K$ of contributing clients per round, federated learning can produce a jointly trained model with high accuracy~\cite{mcmahan2017communication,konevcny2016federated}.
Given that the FedSGD and Fedaveraging are mathematically equivalent, we consider FedSGD aggregation in this paper without loss of generality. 
For privacy-preserving federated learning, the server will provide the differentially private version of the federated model instead of the non-private version for clients to download, together with the set of differential privacy parameters.



\subsection{Problem Statement}
Recent proposals for differentially private federated learning, represented by~\cite{mcmahan2017learning,geyer2017differentially} (as illustrated in Algorithm~\ref{server_dpsgd_12252020}), add Gaussian noise to the per-client shared gradient updates at server each round $t\leq T$. They differ mainly in whether the clipping operation is performed at the server or at each client upon the completion of local training at round $t$. We coin these approaches as Fed-SDP in this paper since they inject noise to the per-client gradient updates at each round $t$. The effect and cost of these clipping and per-client noise injection, whether is done at the server or per client, deliver the same client-level differential privacy guarantee~\cite{mcmahan2017learning}.

\begin{algorithm}[t]
\footnotesize
\caption{\footnotesize Fed-SDP} \label{server_dpsgd_12252020}
\KwIn{$N_i$: \# local data, $D_i$: local data, $K_t$: \# clients per round, $K$: \# total clients, $T$: maximum global round, $\sigma$: noise scale, $C$: clipping bound.} 
 \textbf{Server initialization and broadcast:} global model $W(0)$ of $M$ layers, local batch size $B$, \# local iteration $1\leq L \leq \lceil N_i/B \rceil$, local learning rate $\eta$. \\ 
 \For{round $t$ in $\{0,1,2...T-1\}$}{
  \vspace{0.2cm}

  \textbf{client $i$ does non-private local training and send out local updates:} $\Delta W_i(t) \leftarrow W_i(t)_{L} - W(t)$.  \\

  \vspace{0.2cm}
  \textbf{server do} \\
 \nonl   \textbf{// collect updates from $K_t$ clients} \\
  $\Delta W_{k}(t), k=1,...K_t$ \\  
   \For{$k$ in 1,...,$K_t$}{
   \For{$m$ in $1,...M$ }{ 
    \nonl    \textbf{// compute $L2$ norm for layer $m$} \\
        $||\Delta  W_{ij}(t)_{lm}||_2$  \\
 \nonl         \textbf{// clip per-client update for layer $m$} 
         \begin{align*}
        \vspace{-0.5cm}
         \overline \Delta W_{k}(t)_m &\leftarrow 
        = \Delta W_{k}(t)_m/\max(1, \frac{||\nabla  W_{k}(t)_m||_2}{C}) \vspace{-0.5cm}  \end{align*}     \\
        }
         \nonl   \textbf{// obtain the clipped per-client updates}  \\
         $\overline \Delta W_{k}(t) \leftarrow \{\overline \Delta W_{k}(t)_{m}\} , m=1,..,M$
         }
\nonl \textbf{// compute sanitized updates with $ S  \leftarrow  C $} \\ 
  $\widetilde \Delta W(t) \leftarrow \frac{1}{K_t}(\sum\nolimits_{k = 1}^{{K_t}}  (\overline \Delta W_{k}(t) + \mathcal{N}(0, {\sigma^2 S^2}))$ \\
 \nonl  \textbf{// aggregation:}\\
  $W(t + 1) \leftarrow  W(t) + \frac{1}{K_t}\sum\nolimits_{k = 1}^{{K_t}} \widetilde \Delta w_{k}(t) $\\
 }
 \textbf{Output:} global model $W(T)$
 \end{algorithm}

Recall the threat model in Section~\ref{sec3}, the Fed-SDP approach is resilient to both type 0\&1 gradient leakage attacks if differential privacy is performed by adding Gaussian noise at each client and the adversary intercepts the noisy per-client local parameter updates. The  noisy per-client gradient updates make the leakage attack much harder to succeed due to poor reconstruction inference, especially as the number of rounds increases in federated learning.
Of course, for type-1 gradient leakage, if the Gaussian noise is added at the server but the type-1 leakage is performed on the true local gradient updates at a client, then this version of the Fed-SDP is only resilient to type-0 leakage and cannot protect against type-1 leakage. However, existing versions of Fed-SDP are vulnerable against type-2 leakages which steal per-example gradients during local training, because simply adding Gaussian noise to the per-client local parameter update upon completion of the local training at each round $t$ will fail miserably against the per-example type-2 leakages. This motivates the design of our Fed-CDP to provide a per-example based client differential privacy guarantee in privacy-preserving federated learning.

\begin{figure}[t]
\centerline{\includegraphics[scale=.50]{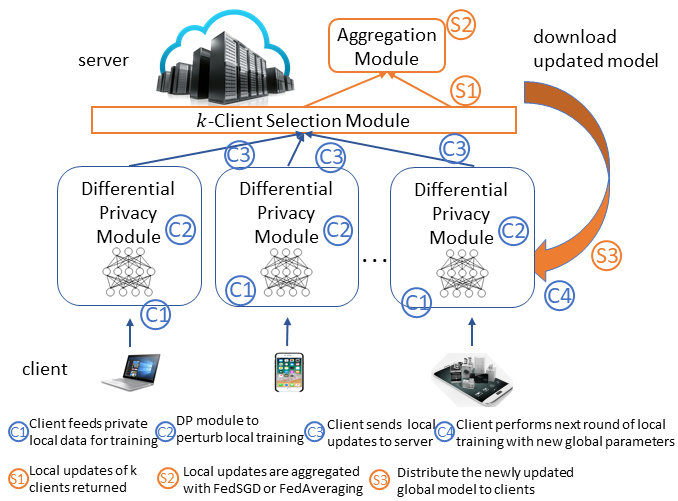}}
\caption{\small System schema of Fed-CDP}
\label{figure:system_schema}
\vspace{-0.4cm}
\end{figure}

\begin{algorithm}
\footnotesize
\caption{\footnotesize Fed-CDP}\label{client_dpsgd_12252020}
\KwIn{$N_i$: \# local data, $D_i$: local data, $K_t$: \# clients per round, $K$: \# total clients, $T$: maximum global round, $\sigma$: noise scale, $C$: clipping bound} 
\textbf{Server initialization and broadcast:} global model $W(0)$ of $M$ layers, local batch size $B$, \# local iteration $1\leq L \leq \lceil N_i/B \rceil$, local learning rate $\eta$. \\ 
\For{round $t$ in $\{0,1,2...T-1\}$}{
 \vspace{0.2cm}

 \textbf{client $i$ do} \\
   \nonl  \textbf{// download current global model:}\\
    $W_i(t)_0 \leftarrow W(t)$\\
    \For{ local iteration $l$ in $\{0,1,2...L-1\}$}{
      \nonl \textbf{// batch processing, take subset of $D_i$} \\
       \For{$j$ in $\{1,...B\}$ }{ 
         \For{$m$ in $\{1,...M\}$ }{ 
      \nonl  \textbf{// compute per example gradient for layer $m$} \\
        $\nabla W_{ij}(t)_l\leftarrow \{\nabla W_{ij}(t)_{lm}\}$ \\
      \nonl    \textbf{// compute $L2$ norm for layer $m$} \\
        $||\nabla  W_{ij}(t)_{lm}||_2$ \\
      \nonl   \textbf{// clip per-example gradients for layer $m$}  \\
        $\overline \nabla W_{ij}(t)_{lm} \leftarrow \nabla W_{ij}(t)_{lm}/\max(1, \frac{||\nabla  W_{ij}(t)_{lm}||_2}{C})$ 
         }
       \nonl   \textbf{// obtain the clipped per-example gradients}  \\
         $\overline \nabla W_{ij}(t)_l \leftarrow \{\overline \nabla W_{ij}(t)_{lm}\} , m=1,..,M$
     }
 \nonl \textbf{// compute sanitized batch gradients with $ S  \leftarrow  C $} \\ 
 \vspace{-0.4cm}
         \begin{equation*}  \widetilde \nabla W_i(t)_{l} \leftarrow \frac{1}{B}\sum \nolimits_{j=1}^B (\overline \nabla W_{ij}(t)_l + \mathcal{N}(0, \sigma^2 S^2))  
 \end{equation*}   \vspace{-0.2cm}\\
 \nonl  \textbf{// gradient descent:} \\
 $W_i(t)_{l+1}  \leftarrow W_i(t)_{l}  - \eta \widetilde \nabla w_i(t)_l$ \\
 } 
 \nonl    \textbf{// send out local updates:} \\
$ \Delta W_{i}(t) \leftarrow W_i(t)_{L} - W(t) $  // with $M$ layers \\ 
  \vspace{0.2cm}
 \textbf{server do} \\
  \nonl \textbf{// collect local updates from $K_t$ clients} \\
  $\Delta W_{i}(t), i=1,...K_t$ \\  
  \nonl \textbf{// aggregation:}\\
 $W(t + 1) \leftarrow  W(t) + \frac{1}{K_t}\sum\nolimits_{i = 1}^{{K_t}} \Delta w_{i}(t) $
} 
\textbf{Output:} global model $W(T)$
\end{algorithm}

\subsection{Fed-CDP: Per-example Client Differential Privacy}


Figure~\ref{figure:system_schema} gives a system overview of Fed-CDP.
Fed-CDP provides gradient leakage resilience against all three types of attacks by adding DP-guided Gaussian noise to the per-example gradients once they are computed before local batch averaging in each of the $L$ iterations during the local training at a client. This enables the support for instance-level differential privacy guarantee. Algorithm~\ref{client_dpsgd_12252020} gives the pseudo-code for Fed-CDP. We first discuss a set of important differential privacy parameters used in Fed-CDP.

{\bf Random Subsampling.\/} Given a total of $K$ participants (clients) and a total of T rounds for distributed learning, in each round t ($1 \leq t \leq  T$), a subset of $K_t$ clients is randomly sampled with replacement under probability $q$. The random sampling by selecting each client independently with probability $q$ is essential for providing low privacy loss through tracking privacy spending using the moments accountant~\cite{abadi2016deep}. 

{\bf Choosing Clipping Strategy $C$.\/} When clipping the client gradient updates, a tradeoff should be made. To keep the noise variance small, we should choose a small clipping upper bound $C$. However, if the goal is to maintain the original update values as much as possible, a too-small clipping bound  $C$ should not be chosen. Instead, we can use the median norm of all original updates from the training set of the client $i$ and use this median as the clipping bound, i.e., $C\leq$ median$_{i\in N_i} \nabla w_i(t)$. Alternatively, we can also define clipping as a function of learning rate $\eta$. With adaptive learning rate methods, in which the learning rate is systematically decreasing as the model converges~\cite{wu2019demystifying}. 


{\bf Clipping Bound $C$ and Sensitivity $S$.\/} To provide a fair comparison with Fed-SDP, in this paper we follow~\cite{abadi2016deep}: which uses a preset fixed clipping bound $C$ to confine the influence of the local model update under $L2$ norm and also use the clipping bound $C$ to estimate the sensitivity for adding noise in Gaussian Mechanism (recall Definition 2). A trade-off should be considered when clipping the contribution of the per-example gradient, as the clipping bound can affect both the noise variance and the informative contribution of the gradient. A large clipping bound would significantly enlarge the Gaussian noise variance and thus pushing the gradient away from its training goal and may thus downgrade or even fail the training, since the large gradient reflects an aggressive step of modification on the model weight for gradient descent. On the other hand, a small clipping bound may keep noise variance stay small but may end up pruning those informative gradients. Empirically,  following the literature~\cite{abadi2016deep}, $C=4$ is used as the default unless otherwise stated.

{\bf Choosing noise scale $\sigma$.\/} By Definition~2, $\sigma$ controls the noise level and by moment accounting in Definition 5 (Equation~\ref{equa:momentsaccountant}), with fixed $\delta$, the privacy loss accumulates inverse-proportionally as noise scale $\sigma$ changes. Existing literature selects the noise scale based on the learning method and task. For example, \cite{geyer2017differentially} uses a small noise scale $\sigma=1$ with clipping bound $C=1$ for client-level differential privacy and \cite{mcmahan2017learning} uses $\sigma$ as small as 0.012 with clipping bound $C=15$ tailor to the local model update in the federated language model. \cite{abadi2016deep} sets its default $\sigma$ to 6 with clipping $C=4$ for privacy accounting via moments accountant. \cite{yu2019differentially} considered a large $\sigma$ of $10$ with clipping $C=4$ for centralized differentially private model publishing, in which the batch gradient is less sensitive to noise. In this paper, we adopt the same setting as \cite{abadi2016deep}: $\delta=1e-5$, $\sigma=6$ with $C=4$ as the default unless otherwise stated.


{\bf Workflow of Fed-CDP algorithm.\/} In each round $t\leq T$ of federated learning, for each client $i$ from the $K_t$ chosen clients, the local model training is performed on each iteration $l$ over the $L$ local iterations ($1\leq L\leq \lceil N_i/B \rceil$). For each iteration $l$, we compute the per-example gradient layer by layer over the $M$ layers for each instance $j$ in a batch $B$ by minimizing the empirical loss function (Lines 6-9). For each layer $l$, the number of gradients is the same as the number of trainable parameters and there exists an $L2$-norm for the layer accordingly. Therefore, a $M$ layer neural network will have $M$ $L2$ norms, one for each layer. Once the per-example gradient is computed for layer $m$ ($1\leq m \leq M$), we clip the gradient in $L_2$ norm (Line 10), i.e., the $L_2$ norm of layer-wise per-example gradient vector $\nabla w_{ij}(t)_{lm}$ is clipped down to $C$ if $||\nabla w_{ij}(t)_{lm}||_2>C$ and $\nabla w_{ij}(t)_{lm}$ is preserved if $||\nabla w_{ij}(t)_{lm}||_2 \leq C$. The per-example gradient will consist of $M$ components (Line 12). Then Gaussian noise is added to the clipped per-example gradient (Line 14). For each batch, the sanitized per-example gradients are gathered for batch averaging and local stochastic gradient descent (Line 15). Upon the completion of the local training over the $L$ iterations, the local training model update is produced by the client $i$ using the Fed-CDP client module and this noisy local gradient update will be shared with  the server (Line 17) for FedSGD aggregation and global model update for the next round of training (Lines 18-20). It is important to note that any change in the gradient of one layer after it is computed will not change the computation of the gradient of its previous layer. This indicates that clipping and adding noise to computed gradients of one layer cannot protect other layers in the model. Hence, Fed-CDP requires the clipping and sanitization to be performed on the gradient of every layer for all $M$ layers and at every local iteration to ensure that the local model training and the global model from the joint training satisfy the per-example based client differential privacy.

By applying clipping and adding noise layer by layer with back-propagation computed from the loss function to the weights of the first model layer, this ensures that sanitizing of per-example per-layer gradients is done immediately after gradients of one layer is computed, enabling Fed-CDP to provide instant protection of the per-example local training gradients against type-2 leakage at each client. Furthermore, the per-example clipping and adding noise at each local client will have a global effect through the iterative joint training of $T$ rounds, enabling Fed-CDP to be resilient against type-0 and type-1 leakages as well.

\section{Privacy Analysis}

\label{sec5}

The design of Fed-CDP aims to provide differential privacy guarantee at both instance level and client level. In this section, we formally analyze the privacy guarantee by Fed-CDP.

{\bf Privacy at instance-level.\/}  
In the definition of ($\epsilon, \delta$) differential privacy~\cite{dwork2014algorithmic}, the two adjacent inputs $A \subseteq \mathcal{D}$ and $A' \subseteq \mathcal{D}$ differing by only a single example, and the privacy protection by the($\epsilon, \delta$)-differential privacy is to introduce a randomized mechanism such that the differentially private version of the learning algorithm will be used to ensure the privacy at instance level, that is, even if an adversary can observe the output of the algorithm, he cannot tell whether the user information defined by a single training example was used in the computation of the algorithm.

Given a distributed collection $\mathcal{D}$ of training data residing on a total of $K$ disseminated clients, Fed-CDP enforces per-example gradient clipping and sanitization via Gaussian noise injection, leading to instant protection of the per-example local training gradients against type-2 leakage at each client. By Definition~5, for a mechanism $\mathcal{M}$, given the privacy parameters $\delta$ and $\sigma$ ($>0$), $\mathcal{M}$ is $(\epsilon,\delta)$-differentially private if the sample rate $q$ satisfies $q<\frac{1}{16\sigma}$ and $\epsilon$ satisfies the following equation:
\begin{equation}
c2\frac{q\sqrt{T\log(1/\delta)}}{\sigma}\leq  \epsilon < c1q^2T.
\label{equa:momentsaccountant-epsilon}
\end{equation}
In order to leverage the moment accountant~\cite{abadi2016deep} to compute the privacy budget spending ($\epsilon$) of Fed-CDP during the joint training for the $T$ rounds, we simply need to prove that local sampling with replacement across different clients in federated learning can be modeled  as global data sampling with replacement. Thus, federated learning can be viewed as an alternative implementation of centralized training in terms of algorithmic logic.




\begin{prop}
The probability of any class being selected globally is the same as the probability of that class in the global data distribution.
\label{prop:sampling}
\end{prop}

One way to simplify the proof of this proposition is to consider each client hold data points from two classes out of the total number of classes ($Z$), and compute the probability of sampling data points from the two classes from a random chosen client $i$ and the probability of the event of a client sampling data from one of its classes. Based on these two probabilities, we can then compute the probability of a specific class being selected randomly, which would be equivalent to $\frac{1}{Z}$. This indicates that the local sampling with replacement can be modeled as global data sampling with replacement. For Fed-CDP with the total amount of training data $N$ over the total $K$ clients, 
the sampled local data over $K_t$ out of $K$ clients in each round $t$ can be viewed as a global data sampling with sampling size of $B*K_t$ and the sampling rate $\frac{B*K_t}{N}$. Hence, we can compute the privacy spending using moments accountant similar to centralized differential privacy. Hence the locally added per-example noise would have a global privacy effect over the global data.

\textbf{Privacy at client-level} In Fed-CDP, for each round $t$, consider any client $i$ from the chosen $K_t$ clients, the local training parameter updates shared by client $i$ to the server are the local aggregation of noisy per-example gradients over $L$ local iterations with batch size $B$ in each iteration. Thus, the proposed Fed-CDP offers per-example based client differential privacy and at the same time offers joint differential privacy~\cite{kearns2014mechanism} from a per-client perspective. Joint differential privacy requires that for two adjacent inputs K and K$'$, differing by only the update from one client $i$ at each round $t$ of the joint training, the privacy protection by the ($\epsilon, \delta$)-joint differential privacy is to introduce a randomized mechanism such that the differentially private version of the learning algorithm will be used to ensure the privacy at client level: even if an adversary can observe the output of the algorithm, he cannot tell whether the local training parameter updates from a single client was used in computing the output of the algorithm.

\begin{definition}
\textbf{Joint Differential privacy~\cite{kearns2014mechanism}}: A mechanism  $\mathcal{M}$: $\mathcal{D}(K) \rightarrow \mathcal{R}$ satisfies joint ($\epsilon,\delta$)-differential privacy if for any two sets of agents $K$ or $K'$, each with its data as $\mathcal{D}(K), \mathcal{D}(K') \subseteq \mathcal{D}$, the following condition holds.
\begin{equation}
\Pr(\mathcal{M}(\mathcal{D})(K) \in \mathcal{R}) \leq e^{\epsilon}
\Pr(\mathcal{M}(\mathcal{D})(K') \in \mathcal{R}) + \delta.
\label{equa:jdp}
\end{equation}
\end{definition}

Joint differential privacy is a relaxation of the standard differential privacy but yet strong: it implies that for any client $i$ in federated learning (agent), even if all the other clients (agents) except client $i$ collude and share their information, they would not be able to learn about client $i$'s private information~\cite{kearns2014mechanism}. Fed-CDP achieves joint differential privacy by utilizing the Billboard Lemma (lemma~\ref{lemma:billboard}) and considering the differentially private global model at the server shared with all clients as the public information on the Billboard.

\begin{lemma}
\textbf{Billboard Lemma~\cite{wu2017data}} Suppose $\mathcal{M}: \mathcal{D} \rightarrow \mathcal{R}$ is $(\epsilon,\delta)$-differentially private. Consider any set of functions $f_k: \mathcal{D}_i \times \mathcal{R} \rightarrow \mathcal{R'} $, where $\mathcal{D}_i \subseteq  \mathcal{D}$ containing agent $i$'s data. The composition $\{f_i(\prod \nolimits_i  \mathcal{D},  \mathcal{M}( \mathcal{D}))\}$ is $(\epsilon,\delta)$-joint differentially private, where $\prod \nolimits_i:  \mathcal{D} \rightarrow  \mathcal{D}_i $ is the projection to agent $i$'s data.
\label{lemma:billboard}
\end{lemma}



Given that existing Fed-SDP approaches~\cite{mcmahan2017learning,geyer2017differentially} clip and add noise at each round $t$ to the local model update shared by each of the $K_t$ clients, Fed-SDP does not explicitly provide the instance-level client different privacy but it does provides the client-level different privacy guarantee in the sense that for two possible sets of clients, K and K’, differing by only one client $i$ at each round $t$ of the joint training, by introducing a randomized mechanism such that the differentially private version of the learning algorithm will be used to ensure that even if an adversary can observe the output of the algorithm, he cannot tell whether the local training parameter updates from a single client was used in the computing the output of the algorithm. Thus, the ($\epsilon, \delta$)-differential privacy by Fed-SDP provides privacy protection at client level. Note that LDP with federated learning~\cite{truex2020ldp}, which independently protects single value of clients' data locally and does not offer differential privacy guarantee on the globally trained model, is beyond the scope of this paper. 


\section{Privacy-utility Trade-off Analysis}
\label{sec6}

The privacy and utility trade-off is known to be an open challenge~\cite{li2009tradeoff,wang2017privacy}. In Fed-CDP, to provide per-example based client differential privacy, in each round $t$, for each of the selected $K_t$ clients, we clip and then add noise to the per-example gradient before performing the local batch stochastic gradient descent for each of the $L$ iterations in the local model training by the client. The same process of clipping and noise injection repeats for a total of $T$ rounds to produce the global model as the outcome of the joint training.  

We first analyze how the injected Gaussian noise affects the privacy-performance trade-off in terms of changing gradient descent directions and provide a formal guarantee of the maximal distortion which a neural network model can tolerate without deteriorating the training performance.

Let $g_j(x; w(t)): \mathcal{R}^D \rightarrow R$ be the loss function of the continuously differentiable DNN model on class $j$ given the
input $x$. We define $s(x,w(t)) = g_y(x; w(t)) - g_{j,j\neq y}(x; w(t))$ where $y$ is the label of input $x$. Under the assumption of Lipschitz condition, we have 
\begin{equation}
    |s(x,w(t)) -s(x,w'(t))| \leq L_v ||w(t)-w'(t)||_u,
    \label{equa:theoryeq1}
\end{equation}
where $w'(t)$ is the differentially private version of the model parameters $w(t)$,  and $L_v =\max \nolimits_x \{||\nabla{s(x,w(t))}||_v\}$ is the Lipschitz constant with $\frac{1}{u} + \frac{1}{v}=1$ where $u\geq 1$ and $v \leq \infty$. 
By line 14 of Algorithm~\ref{client_dpsgd_12252020}, we can formulate the private model parameters as $w'(t)=w(t) + \xi$ and substitute  Equation~\ref{equa:theoryeq1} as
\begin{equation}
    s(x,w(t)) - L_v||\xi||_u \leq   s(x,w(t)+\xi) \leq   s(x,w(t)) + L_v||\xi||_u.
    \label{equa:theoryeq2}
\end{equation}

Equation~\ref{equa:theoryeq2} shows that when $s(x,w(t)+\xi)<0$, we will have $ g_y(x; w(t)+\xi) < g_{j,j\neq y}(x; w(t)+\xi)$, indicating an alternation of ground-truth class used to compute the loss function. This shows that 
adding large Gaussian noise to the model, may move the loss function computed on the ground-truth class to other classes and lead to the loss of utility with respect to learning convergence and prediction accuracy. Hence, we want the privacy-preserving cost $||\xi||_p$ to be small enough such that $s(x,w(t)) - L_v||\xi||_u \geq 0$.

 \begin{figure}[t]
\centerline{\includegraphics[scale=.58]{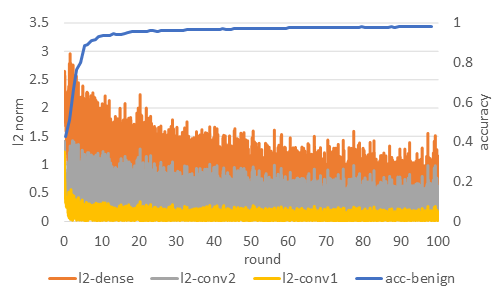}}
\caption{\small Changing of gradients' $L2$ norm during training}
\label{fig:l2norm}
\vspace{-0.4cm}
\end{figure}

\begin{prop} 
\textbf{Upper bound $\xi_u$ for Utility-Privacy trade-off}\\
To preserve the training against the given ground-truth label $y$ and maintain the training information, the maximum change made to model parameters (or gradient) need to be bounded as follows: 
\vspace{-0.3cm}
\begin{equation}
    ||\xi||_u \leq \min_{j \neq y}\frac{g_y(x; w(t)) - g_{j,j\neq y}(x; w(t))}{L_v}.
    \label{equa:theoryeq3}
\end{equation}
\label{theorem:tradeoff}
\vspace{-0.5cm}
\end{prop}


This proposition states that $||\xi||_u \leq \min_{j \neq y}\frac{g_y(x; w(t)) - g_{j,j\neq y}(x; w(t))}{L_v}$ is the lower bound for minimum distortion of the model parameter that would result in the loss of training performance. We can prove the proposition based on the Lipschitz assumption as follows: First, we have 
\begin{align}
& |s(x,w(t)) -s(x,w'(t))| \leq L_v ||w(t)-w'(t)||_u = L_v||\xi||_u \Rightarrow \nonumber \\
& s(x,w(t)) - L_v||\xi||_u \leq    s(x,w(t)+\xi) 
 \leq   s(x,w(t)) + L_v||\xi||_u \label{equa:theoryeq2_repeat}
 \end{align}
To ensure that the distortion does not affect the training process by alternating the prediction towards the ground-truth class to other classes,  we will need $s(x,w(t)) - L_v||\xi||_u \geq 0$ in Equation~\ref{equa:theoryeq2_repeat} holds. Then we have
\begin{align*}
& s(x,w(t)) - L_v||\xi||_u \geq 0 \Rightarrow ||\xi||_u \leq \frac{s(x,w(t))}{L_v}  \\
& \Rightarrow ||\xi||_u \leq \min_{j \neq y}\frac{g_y(x; w(t)) - g_{j,j\neq y}(x; w(t))}{L_v}
 \end{align*}

\noindent Proposition~\ref{theorem:tradeoff} indicates that the training performance will less likely to be affected by the injected Gaussian noise if $||\mathcal{N}(0, \sigma^2 S^2)||_u\leq \min_{j \neq y}\frac{g_y(x; w(t)) - g_{j,j\neq y}(x; w(t))}{L_v}$. 

Given that it is computationally expensive to find such a noise bound in practice, especially for deep learning algorithms with large number of classes. One approach is to translate loss distance between the label class and the second-largest prediction class into the $L2$ magnitude of the gradients and ensure the bound over such $L2$ norm of local gradient updates. 

Figure~\ref{fig:l2norm} illustrates an example of the decaying magnitude of $L2$ norm of the gradient during training. Mean $L2$ norm of 100 MNIST clients at one local iteration is reported. The decaying magnitude of gradient's $L2$ norm indicates that the loss value between the predicted label and ground-truth label decreases over time, implying a decaying confidence gap between the label class $g_y(x; w(t))$ and the class with the second-largest confidence. This is aligned with the observation that gradients in early iterations are more informative and tend to have a larger $L2$-norm value~\cite{wei2020framework}. This is consistent with the literature on the need for fine-tuning of clipping bound used in a differentially private algorithm~\cite{abadi2016deep} to bound the distortion noise using a proper clipping bound with constant clipping method.


{\bf Fed-CDP (decay).\/} Motivated by the utility-privacy tradeoff analysis, we consider to optimize Fed-CDP by using a dynamic decaying clipping method with the initial clipping bound $C$ in order to keep track of the decaying $L2$ sensitivity of gradients as federated learning progresses in the number of rounds, ensuring the differential privacy based distortion noise being bounded properly by following the proposition~\ref{theorem:tradeoff}. It is important to note that decaying clipping bound will not modify the privacy composition. Also the noise variance is determined by both the sensitivity $S$ (thus the the clipping bound $C$ in Fed-CDP and most existing DP algorithms) and the noise scale $\sigma$. Using a dynamic clipping method with decay function on bound $C$ will result in decreasing trend of noise variance over the $T$ rounds of federated learning, resulting in non-uniform distribution of privacy spending over the $T$ rounds according to Lemma~\ref{lemma:gaussian_mechanism}.

Empirical study in Section~\ref{sec7} shows that by decaying clipping bound to track the decaying noise scale, Fed-CDP (decay) further optimizes Fed-CDP with improved accuracy performance and higher resilience to gradient leakage attacks with the same privacy budget spending, achieving better utility-privacy tradeoff, compared to Fed-CDP baseline.

\section{Experimental Evaluation}

\label{sec7}

\begin{table}[t]
\centering
\scalebox{0.85}{
\small{
\begin{tabular}{|c|c|c|c|c|c|}
\hline
                   & MNIST  & CIFAR10 & LFW     & adult  & cancer \\ \hline
\# training data      & 60000  & 50000   & 2267    & 36631  & 426    \\ \hline
\# validation data       & 10000  & 10000   & 756     & 12211  & 143    \\ \hline
\# features        & 28*28  & 32*32*3 & 32*32*3 & 105    & 30     \\ \hline
\# classes         & 10     & 10      & 62      & 2      & 2      \\ \hline
\# data/client     & 500    & 400     & 300     & 300    & 400    \\ \hline
\# local iteration $L$ & 100    & 100     & 100     & 100    & 100    \\ \hline
local batch size $B$   & 5      & 4       & 3       & 3      & 4      \\ \hline
\# rounds  $T$        & 100    & 100     & 60      & 10     & 3      \\ \hline
non-private acc    & 0.9798 & 0.674   & 0.695   & 0.8424 & 0.993  \\ \hline
non-private cost(ms)    & 6.8 & 32.5   & 30.9   & 5.1 & 4.9  \\ \hline

\end{tabular}
}}
\caption{\small Benchmark datasets and parameters}
\label{table:dataset_setup}
\vspace{-0.4cm}
\end{table}

\begin{table*}[t]
\centering
\scalebox{0.99}{
\small{
\begin{tabular}{|c|c|c|c|c|c|c|c|c|c|c|c|c|}
\hline
\multirow{2}{*}{} & \multicolumn{4}{c|}{K=100}     & \multicolumn{4}{c|}{K=1000}     & \multicolumn{4}{c|}{K=10000}   \\ \cline{2-13} 
     Kt/K             & 5\%   & 10\%  & 20\%  & 50\%  & 5\%   & 10\%  & 20\%  & 50\%  & 5\%   & 10\%  & 20\%  & 50\%  \\ \hline
non-private            & 0.924 & 0.954 & 0.959 & \textbf{0.965} & 0.977 & \textbf{0.980} & 0.978 & 0.978 & 0.979 & \textbf{0.980} & \textbf{0.980} & 0.979 \\ \hline
Fed-SDP         & 0.803 & 0.823 & 0.834 & \textbf{0.872} & 0.925 & 0.928 & 0.934 & \textbf{0.937} & 0.935 & 0.939 & 0.941 & \textbf{0.944} \\ \hline
Fed-CDP         & 0.815 & 0.831 & 0.858 & \textbf{0.903} & 0.951 & 0.956 & 0.96  & \textbf{0.964} & 0.966 & 0.963 & \textbf{0.968} & 0.966 \\ \hline
Fed-CDP(decay)  & 0.833  & 0.842  & 0.866  & \textbf{0.909} & 0.968 & 0.975 & \textbf{0.977}  & 0.976 & 0.974 & 0.978 & 0.979 & \textbf{0.980} \\ \hline

\end{tabular}
}}
\caption{\small Accuracy comparison by varying \# total clients and $Kt/K$ in MNIST with default $C=4$ and $\sigma=6$}
\label{table:kt_vary}
\vspace{-0.4cm}
\end{table*} 

\begin{table}[ht]
\centering
\scalebox{0.95}{
\small{
\begin{tabular}{|c|c|c|c|c|c|}
\hline
          & MNIST  & CIFAR-10 & LFW    & adult  & cancer \\ \hline
non-private    & 6.8 & 32.5   & 30.9 & 5.1 & 5.1 \\ \hline
Fed-SDP  & 6.9 & 33.8   & 31.3 & 5.2 & 5.1 \\ \hline
Fed-CDP  & 22.4 & 131.5   & 112.4 & 11.8 & 11.9 \\ \hline
Fed-CDP(decay)  & 22.6 & 132.1 & 114.6  &  12.1 & 12.0  \\ \hline
\end{tabular}
}}
\caption{\small Time cost per local iteration per client(ms)}
\label{table:timecost}
\vspace{-0.2cm}
\end{table}

\begin{table}[ht]
\centering
\scalebox{0.95}{
\small{
\begin{tabular}{|c|c|c|c|c|c|c|}
\hline
         & C=0.5 & C=1   & C=2   & C=4   & C=6   & C=8    \\ \hline
MNIST    & 0.914 & 0.934 & 0.943 & \textbf{0.949} & 0.933 & 0.923   \\ \hline
CIFAR-10 & 0.408 & 0.568 & 0.602 & \textbf{0.633} & 0.624 & 0.611   \\ \hline
LFW      & 0.582 & 0.594 & 0.619 & \textbf{0.649} & 0.627 & 0.601   \\ \hline
adult    & 0.81  & 0.822 & \textbf{0.825} & 0.824 & 0.807 & 0.796   \\ \hline
cancer   & 0.965 & 0.972 & \textbf{0.979} & \textbf{0.979} & 0.972 & 0.972   \\ \hline
\end{tabular}
}}
\caption{\small  Accuracy of Fed-CDP by varying clipping with default $\sigma=6$}
\label{table:clipping_vary}
\vspace{-0.4cm}
\end{table}

  \begin{table}[ht]
\centering
\scalebox{0.95}{
\small{
\begin{tabular}{|c|c|c|c|c|c|c|}
\hline
         & $\sigma$=0.5 & $\sigma$=1 & $\sigma$=2 & $\sigma$=4 & $\sigma$=6 & $\sigma$=8  \\ \hline
MNIST    & 0.956        & 0.954      & 0.952      & 0.951      & 0.949      & 0.934      \\ \hline
CIFAR-10 & 0.646        & 0.641      & 0.639      & 0.634      & 0.633      & 0.612       \\ \hline
LFW      & 0.683        & 0.678      & 0.672      & 0.667      & 0.649      & 0.646       \\ \hline
adult    & 0.838        & 0.837      & 0.836      & 0.834      & 0.824      & 0.822    \\ \hline
cancer   & 0.993        & 0.993      & 0.993      & 0.993      & 0.979      & 0.979     \\ \hline
\end{tabular}
}}
\caption{\small Accuracy of Fed-CDP by varying noise scale with default $C=4$}
\label{table:noise_scale_vary}
\vspace{-0.2cm}
\end{table}

We evaluate Fed-CDP on five benchmark datasets as shown in Table~\ref{table:dataset_setup}, with 
each dataset given the parameter setup, the validation accuracy and training cost for the non-private federated learning model. 
MNIST is a grey-scale hand-written digits dataset consisting of 60,000 training data and 10,000 test data.  
The 5:1 train-validation ratio is applied to the 60,000 training images. We partition the 50,000 training data into shards. Each client gets two shards with 500 samples from two classes. For CIFAR10, the 4:1 train-validation ratio is applied to the 50,000 training images. Each client has 400 images from two classes. Labeled Faces in the Wild(LFW) dataset has 13233 images from 5749 classes. The original image size is $250\times 250$ and we crop it into $32\times 32$ size to extract the 'interesting' region. Since most of the classes have a very limited number of data points, we consider the 62 classes that have more than 20 images per class. For a total number of 3023 eligible LFW data, a 3:1 train-validation ratio is applied. We partition the training data into shards and each client has 300 images from about 15 classes. We evaluate the three benchmark image datasets on a multi-layer convolutional neural network with two convolutional layers and one fully-connected layer. For the two attribute datasets, the 3:1 train-validation ratio is applied to both. {\bf Adult income} dataset has 48842 total training data with 105 converted numerical features, and each client holds 300 data from two classes randomly sampled from the training data. {\bf Breast cancer} dataset has 569 training data with 30 numerical features, and 
each client has a full copy of the dataset due to its small number of data points. A fully-connected model with two hidden layers is used for attribute datasets.

Our federated learning setup follows the simulator in~\cite{geyer2017differentially} with the total $K$ clients varying from 100, 1000 to 10,000, and $K_t$ is set at a varying percentage of $K$, such as $5\%, 10\%, 20\%, 50\%$ per round. 
All experiments are conducted on an Intel 4 core i5-7200U CPU@2.50GHz machine with an Nvidia Geforce 2080Ti GPU.
We evaluate Fed-CDP in terms of three sets of metrics: (i) training performance in validation accuracy and time cost of conducting one local iteration at a client (ms/iteration), (ii) the privacy budget spending $\epsilon$ with varying settings of other differential privacy parameters, such as clipping bound $C$ and noise scale $\sigma$; and (iii) the resilience against all three types of gradient leakage attacks in terms of attack success (Yes or No), \#attack iterations to succeed with the default set to 300, and attack reconstruction distance, defined by the root mean square deviation between reconstructed input $x_{rec}$ and its private ground-truth counterpart $x$: $\frac{1}{A}\sum\nolimits_{i = 1}^A {(x(i) - {x_{rec}}(i)} {)^2}$. $A$ denotes the total number of features in the input.

\subsection{Training performance Evaluation}

\textbf{Fed-SDP v.s. Fed-CDP.} The performance of Fed-CDP is evaluated and compared with Fed-SDP by the default parameter setting of MNIST in Table~\ref{table:dataset_setup}. Table~\ref{table:kt_vary} shows the results by varying total $K$ clients from 100 to 1000 and 10000 with 5\%, 10\%, 20\% and 50\% participating clients at each round. We make two observations: (1) Both Fed-CDP and Fed-SDP can achieve high accuracy comparable to that of the non-private setting.  (2) Fed-CDP consistently achieves higher accuracy than Fed-SDP for all settings of $K$ and $K_t$ while Fed-CDP(decay) consistently improves Fed-CDP with slightly better accuracy performance. For Fed-CDP(decay), we linearly
decay the clipping bound from C=6 to C=2 in 100 rounds and achieve high validation accuracy of 0.9748, on par with the non-private FL setting on MNIST. We will show later that decay-clipping optimized Fed-CDP also demonstrates higher resiliency against gradient leakage attacks.

Table~\ref{table:timecost} shows the per local iteration per client training time comparison.
Given all $K_t$ clients can perform local training in parallel, the overall time spent for one round is determined by the slowest client local training at the round $t$. Although Fed-CDP incurs additional computation overhead to clip and sanitize per-example gradients in each local batch (iteration), compared to Fed-SDP, the absolute time cost  remains relatively low. The decay of clipping bound brings a negligible cost compared to Fed-CDP.

\textbf{Fed-CDP performance with varying parameter settings.} 
We first study the effect of varying clipping bound from 0.5 to 8. Table~\ref{table:clipping_vary} shows the results of Fed-CDP. By combining gradient clipping and noise injection to per-example gradients, Fed-CDP leads to high accuracy on MNIST, Adult and Cancer when $C=2$ or $C=4$ and high accuracy on LFW and CIFAR-10 when $C=4$ or $C=6$, because for different datasets, $L2$ norm of the gradients varies and results in different best clipping bound settings.  Since the clipping bound will impact both the gradient information preservation and noise variance, the highest accuracy appears only with an appropriately set clipping bound and the performance will downgrade if the clipping bound is too large or too small.
We next evaluate the impact of varying noise scale of $0.5, 1, 2, 4, 6, 8$. Table~\ref{table:noise_scale_vary} shows the results given the clipping bound $C=4$. Noise scale controls the amount of Gaussian noise added to the per-example gradients. 
The results confirm that adding too much noise will impact negatively the training performance. In Fed-CDP, we set $\sigma=6$ by default to ensure the requirement of $q<\frac{1}{16\sigma}$ for moments accountant, even though a smaller noise scale may bring improvement on the accuracy performance for all datasets.

\subsection{Privacy Evaluation}

We evaluate and compare the privacy composition of Fed-CDP and Fed-SDP using moments accountant (recall Definition~5 and Section~\ref{sec5}) on all 5 datasets with two settings on the per-client location training iterations: $L=1$ and $L=100$.  Table~\ref{privacy_table} summarizes the results.

\textbf{Privacy at instance level.} 
Since privacy composition is only related with the sampling rate and the total number of rounds ($T$) under a given $\delta$  by Equation~\ref{equa:momentsaccountant}, $\epsilon$ for different dataset depends mainly on the total number of rounds since we set the global sampling rate for all datasets to 0.01. 
The training model for MNIST and CIFAR-10 under $L=100$ is $(0.8227, 1e-5)$-differential private with $\epsilon=0.8227$ and $\delta=1e-5$. For LFW, adult, and cancer datasets, we have $\epsilon =0.6356$,  $\epsilon =0.2761$, and $\epsilon =0.1469$ respectively.  All 5 datasets have the same sampling rate of 0.01, and their total training rounds $T$ set to 100, 100, 60, 10, and 3 for MNIST, CIFAR-10, LFW, adult, and cancer respectively. Note that although we report the instance-level privacy by round since we will only get a new global model for each round, the actual $\epsilon$ value accumulates with the setting of the \# of local iterations ($L$). For $L=1$, clients performing local model training over one local iteration per round, which will generate a differentially private model with a much smaller privacy budget spending $\epsilon$ compared to the $\epsilon$ from the differentially private model generated over clients with local training on 100 local iterations at the same round. However, local training with $L=1$ takes more rounds to achieve the same accuracy as the local training with $L=100$ iterations.



\begin{table}[t]
\centering
\scalebox{0.85}{
\small{
\begin{tabular}{|c|c|c|c|c|c|c|}
\hline
\multicolumn{2}{|c|}{}                                    & MNIST     & CIFAR-10  & LFW      & adult    & cancer   \\ \hline
\multirow{4}{*}{\rotatebox{90}{instance}} & Fed-CDP: 1   & 0.0845  & 0.0845  & 0.0689 & 0.0494 & 0.0467 \\ \cline{2-7} 
                                & Fed-CDP: 100 & 0.8227    & 0.8227    & 0.6356   & 0.2761   & 0.1469   \\ \cline{2-7} 
                                & Fed-SDP: 1    & \multicolumn{5}{c|}{\multirow{2}{*}{not supported}} \\ \cline{2-2}
                                & Fed-SDP: 100 & \multicolumn{5}{c|}{}                                  \\ \hline
\multirow{4}{*}{\rotatebox{90}{client}}   & Fed-CDP: 1  & 0.0845  & 0.0845  & 0.0689 & 0.0494 & 0.0467 \\ \cline{2-7} 
                                & Fed-SDP: 1   & 0.8536    & 0.8536    & 0.6677   & 0.3025   & 0.2065   \\ \cline{2-7} 
                                 & Fed-CDP: 100 & 0.8227    & 0.8227    & 0.6356   & 0.2761   & 0.1469   \\ \cline{2-7} 
                                & Fed-SDP: 100 & 0.8536    & 0.8536    & 0.6677   & 0.3025   & 0.2065   \\ \hline
\end{tabular}
}}
\caption{\small Privacy composition of Fed-SDP and Fed-CDP using $\epsilon$ with $\delta=1e-5$, $L=1$ and $L=100$ local iterations for per-client local training. $\epsilon$ measured at 100, 100, 60, 10, 3 rounds for MNIST, CIFAR10, LFW, adult, and cancer dataset respectively.}
\label{privacy_table}
\vspace{-0.4cm}
\end{table}

\begin{table*}[t]
\centering
\scalebox{0.95}{
\small{
\begin{tabular}{|c|c|c|c|c|c|c|c|c|c|}
\hline
   \multicolumn{2}{|c|}{\multirow{2}{*}{}} & \multicolumn{4}{c|}{type 0\&1}                                                               & \multicolumn{4}{c|}{type 2}                                                                 \\ \cline{3-10} 
\multicolumn{2}{|c|}{}                  & non-private & Fed-SDP & Fed-CDP & \begin{tabular}[c]{@{}c@{}}Fed-CDP \\ (decay)\end{tabular} & non-private & Fed-SDP & Fed-CDP & \begin{tabular}[c]{@{}c@{}}Fed-CDP\\ (decay)\end{tabular} \\ \hline
\multirow{3}{*}{\rotatebox{90}{MNIST}} & succeed & \textcolor{red}{Y}           & N       & N       & N               & \textcolor{red}{Y}           & \textcolor{red}{Y}       & N       & N              \\ \cline{2-10} 
                       & reconstruction distance  & 0.1549      & 0.6991  & 0.7695  & 0.937           & 0.0008      & 0.0008  & 0.739   & 0.943          \\ \cline{2-10} 
                       & attack iteration & 6           & 300     & 300     & 300             & 7           & 7       & 300     & 300            \\ \hline
\multirow{3}{*}{\rotatebox{90}{LFW}}   & succeed & \textcolor{red}{Y}           & N       & N       & N               & \textcolor{red}{Y}           & \textcolor{red}{Y}       & N       & N              \\ \cline{2-10} 
                       & reconstruction distance  & 0.2214      & 0.7352  & 0.8036  & 0.941           & 0.0014      & 0.0014  & 0.6626  & 0.945          \\ \cline{2-10} 
                       &  attack iteration  & 24          & 300     & 300     & 300             & 25          & 25      & 300     & 300            \\ \hline
\end{tabular}
}}
\caption{{\small Average attack effectiveness comparison from 100 clients, with the maximum attack iteration set to 300.}}
\label{table:attack_evaluation}
\vspace{-0.4cm}
\end{table*}

\textbf{Privacy at client level.} 
According to the Billboard Lemma(lemma~\ref{lemma:billboard}), with $\delta=1e-5$, the trained model with Fed-CDP after the $T$ rounds will accumulate the privacy spending $\epsilon=0.8227, 0.8227, 0.6356, 0.2761, 0.1469$ for where for MNIST, CIFAR-10, LFW, adult and cancer respectively where $T=100, 100, 60, 10, 3$ and $L=100$ under joint differential privacy. Privacy at client-level for Fed-CDP benefits from the instance-level privacy for Fed-CDP. For Fed-SDP, the client-level differential privacy is computed with the sampling rate $q2=\frac{K_t}{K}$~\cite{mcmahan2017learning,geyer2017differentially} at each round. Since Fed-SDP adds noise directly to the per-client local gradient updates and whether the per-client local training is performed over one local iteration or 100 local iterations will not affect its privacy accounting, as shown in Table~\ref{privacy_table}.

\subsection{Gradient-Leakage Resiliency}

\begin{figure}[t]
\centerline{\includegraphics[scale=.65]{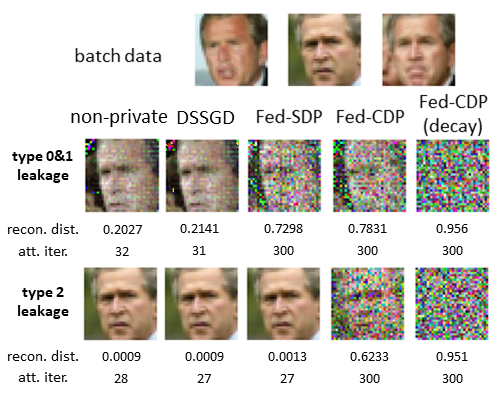}}
\caption{\small Visualization of gradient leakage attacks under different Fed-DP modules. Gradients of one example in the batch is used to do type-2 attack.}
\label{fig:attack_vis}
\vspace{-0.4cm}
\end{figure}

We measure the resiliency of Fed-CDP and Fed-SDP against gradient leakage attacks using MNIST and LFW dataset. The gradient leakage attack experiment is performed on gradients from the first local iteration as gradients at early training iterations tend to leak more information than gradients in the later stage of the training~\cite{wei2020framework}. Table~\ref{table:attack_evaluation} reports the resilience results in terms of attack success (yes or no), the average reconstruction distance, and attack iterations from 100 clients. For failed attacks, we run the attack reconstruction and record the attack reconstruction distance at the maximum attack iterations of 300. We observe that non-private federated learning is vulnerable to all three types of gradient leakage attacks, while Fed-CDP can effectively mitigate all three types of leakages and Fed-SDP is vulnerable to the type-2 leakage though it is more resilient to the type-0 and type-1 leakage compared to non-private FL. 
Compared to Fed-SDP, Fed-CDP has a larger reconstruction distance, which implies more robust masking of the private training data against gradient leakage attacks. 
Figure~\ref{fig:attack_vis} shows a visualization of Fed-CDP and Fed-SDP under three types of gradient leakages using an example batch of LFW dataset in the local training, showing Fed-CDP(decay) offers the highest resilience against all three gradient leakages, followed by Fed-CDP, while Fed-SDP is vulnerable to type-2 leakage and non-private FL and Distributed
Selective SGD(DSSGD~\cite{shokri2015privacy}) is vulnerable to all three types of gradient leakages.

\begin{figure}[t]
 \centerline{\includegraphics[scale=.60]{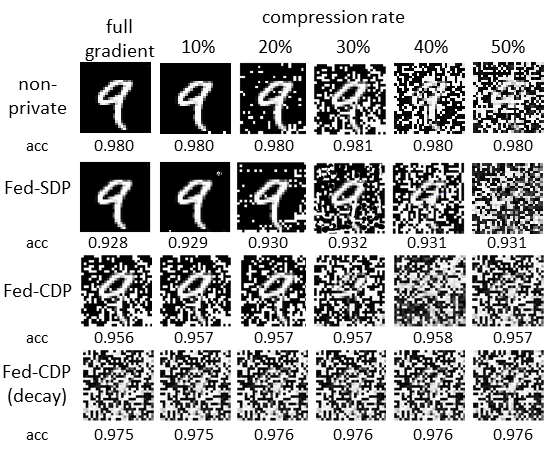}}
 \vspace{-0.2cm}
  \caption{\small Accuracy and resiliency to type-2 leakage in communication efficient federated learning}
  \label{fig:attack_ensemble} 
\vspace{-0.4cm}
\end{figure}

Figure~\ref{fig:attack_ensemble} provides a visualization comparison of Fed-CDP(decay), Fed-CDP, and Fed-SDP under type-2 gradient leakage attack in the communication-efficient federated learning using MNIST dataset. Validation accuracy is also reported at a total of 1000 clients and 100 participants in the figure. In this set of experiments, we consider communication-efficient federated learning by pruning the insignificant gradients, which make little contribution to joint training, i.e., gradients with very small values~\cite{wei2020framework}. Using non-private training or Fed-SDP, the gradients shared via communication-efficient protocol (by compression) may still suffer from the gradient leakage threats for compression ratios up to 30\% on MNIST. In contrast, Fed-CDP is more resilient than Fed-SDP, while Fed-CDP(decay) provides the highest resilience in all settings against the type-2 leakages in the communication-efficient federated learning protocols.


\section{Conclusion}

We have presented a per example based client different privacy approach to federated learning with three original contributions. We first identified three types of gradient leakage attacks and explained why the existing Fed-SDP approach is vulnerable. We then introduce Fed-CDP, the per example-based client differential privacy algorithm, with a formal comparison between Fed-CDP and Fed-SDP in terms of privacy accounting. Finally,  we provide a formal analysis of the privacy-utility trade-off for providing differential privacy guarantee by Fed-CDP and present Fed-CDP(decay), which further improves the accuracy and resiliency of Fed-CDP by incorporating a dynamic decay clipping method, instead of constant clipping method popularly used in the literature~\cite{abadi2016deep,mcmahan2017learning}. With extensive evaluation, we show that the Fed-CDP approach outperforms Fed-SDP with high resilience against all three types of gradient leakage attacks while offering competitive accuracy performance.


\vspace{0.4cm}

\textbf{Acknowledgement.}
This work is partially supported by the National Science Foundation under Grants NSF 1564097 and NSF 2038029 as well as an IBM faculty award. The authors especially thank the support and collaboration with the systems group in IBM T.J. Watson, led by Dr. Donna N Dillenberger. 
Any opinions, findings, and conclusions or recommendations expressed in this material are those of the author(s) and do not necessarily reflect the views of the National Science Foundation or IBM.

\bibliographystyle{IEEEtran}
\bibliography{reference}

\end{document}